\documentclass[runningheads]{llncs}
\usepackage{graphicx}
\usepackage{subcaption}
\usepackage{amsmath}
\usepackage{url}
\usepackage{color}
 \usepackage{xcolor}
\begin{document}
\title{Efficient View Clustering and Selection for City-Scale 3D Reconstruction}
%
%
%
%
%

\author{Marco Orsingher\inst{1,2} \and Paolo Zani\inst{2} \and Paolo Medici \inst{2} \and Massimo Bertozzi \inst{1}}
\institute{Università degli Studi di Parma \\ Dipartimento di Ingegneria e Architettura \and Vislab Srl - Ambarella Inc \\}


\maketitle              
\begin{abstract}
Image datasets have been steadily growing in size, harming the feasibility and efficiency of large-scale 3D reconstruction methods. In this paper, a novel approach for scaling Multi-View Stereo (MVS) algorithms up to arbitrarily large collections of images is proposed. Specifically, the problem of reconstructing the 3D model of an entire city is targeted, starting from a set of videos acquired by a moving vehicle equipped with several high-resolution cameras. Initially, the presented method exploits an approximately uniform distribution of poses and geometry and builds a set of overlapping clusters. Then, an Integer Linear Programming (ILP) problem is formulated for each cluster to select an optimal subset of views that guarantees both visibility and matchability. Finally, local point clouds for each cluster are separately computed and merged. Since clustering is independent from pairwise visibility information, the proposed algorithm runs faster than existing literature and allows for a massive parallelization. Extensive testing on urban data are discussed to show the effectiveness and the scalability of this approach.

\end{abstract}

\section{Introduction}

Nowadays, arbitrarily large datasets of high-resolution images have become available thanks to almost unlimited sources of data, such as videos acquired by autonomous vehicles with multi-camera rigs and Internet pictures uploaded by millions of users on Social Media. While this large amount of data can be used to effectively reconstruct a precise 3D view of the world, the increasing size and high redundancy introduce several challenges for 3D reconstruction. A traditional image-based 3D reconstruction pipeline is composed by two main steps: Structure From Motion (SFM) and  Multi-View Stereo (MVS). SFM algorithms produce camera poses and sparse point clouds with raw images as input~\cite{1_schonberger2016structure}. Then, MVS is used to generate depth and normal maps for each image, which can be converted either to dense point clouds or meshed surfaces~\cite{2_seitz2006comparison}. 

\begin{figure*}
    \centering
    \includegraphics[width=\textwidth]{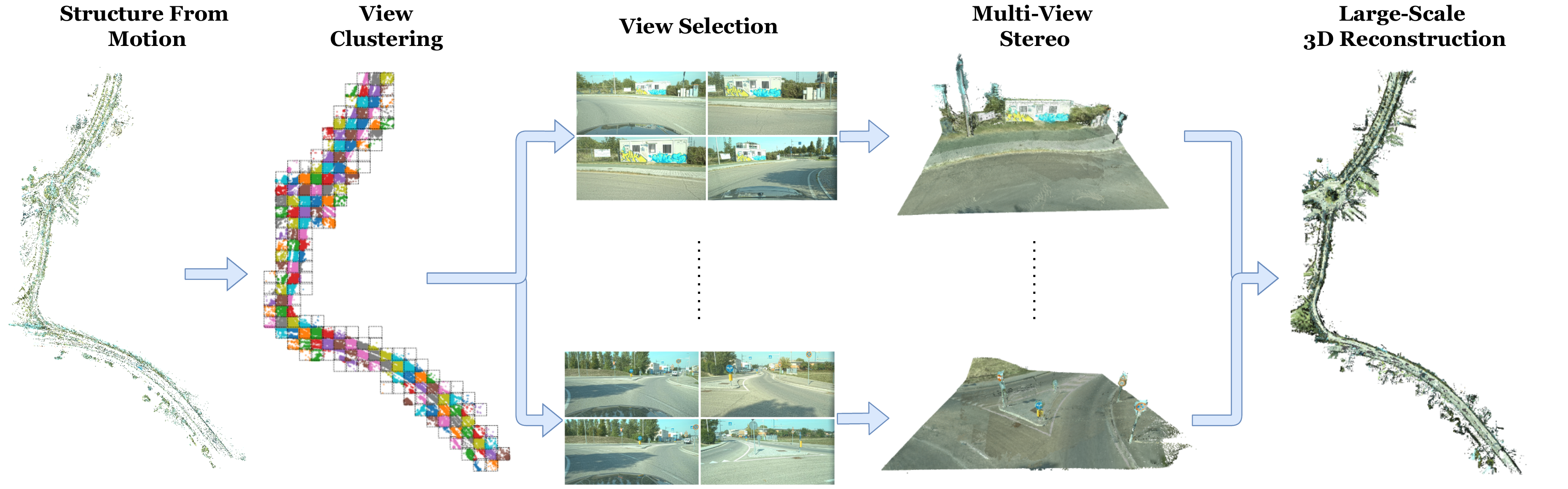}
    \vspace*{1mm}
    \caption{Overview of the proposed approach.}
    \label{fig:overview}
\end{figure*}

To deal with such efficiency issues, view selection algorithms have been proposed in order to select a representative subset of input images~\cite{3_mauro2014integer,4_mauro2014unified}. Moreover, each image shares common visibility information only with a local group of neighboring views, therefore the whole dataset can be divided in partially overlapping clusters to be processed simultaneously and independently~\cite{5_ladikos2009spectral,6_furukawa2010towards,7_mauro2013overlapping,8_zhang2015joint}. Several approaches have been developed to reconstruct large buildings using community photo collections~\cite{9_agarwal2011building,6_furukawa2010towards,8_zhang2015joint}. These works rely on unique and recurrent architectonic features that belong to the building of interest, when computing shared visibility information among images. However, urban scenarios exhibit a peculiar set of challenges and they have rarely been targeted in detail~\cite{10_akbarzadeh2006towards}. The presence of large portions of textureless surfaces, such as roads, vegetation or sky, makes it difficult to extract well-distributed and well-triangulated keypoints during the SFM phase. 

In this paper, a novel framework for efficient view clustering and selection for images of urban scenarios is presented. The proposed approach is specifically devoted to city-scale 3D reconstruction and designed to solve the aforementioned issues, assuming an arbitrarily large dataset of images acquired by a moving vehicle. The method overview is shown in figure~\ref{fig:overview}: starting from the output of a SFM module, the view clustering algorithm builds uniform clusters in the observed world, while the view selection step computes the optimal subset of views in parallel for each cluster. Then, each cluster is reconstructed separately with any MVS method of choice and partial results are merged to get the full 3D model of the scene. 

\section{Related Work\label{sect:related}}

Scalability issues for large-scale 3D reconstruction mainly arise in literature in the context of architectural datasets collected from unstructured Internet images. Furukawa \textit{et al.}~\cite{6_furukawa2010towards} propose to perform global view selection on the whole set of images to remove redundancy and to build a visibility graph with the remaining cameras. These similarities are collected in a matrix that represents the adjacency matrix of the visibility graph. Then, an optimization procedure is applied to iteratively divide the graph into clusters using normalized-cuts, enforcing a size constraint, and add cameras back if a coverage constraint is violated. This process is repeated until convergence. 

Several other approaches are based on this visibility graph formulation. Ladikos \textit{et al.}~\cite{5_ladikos2009spectral}  apply spectral graph theory on the similarity matrix and use mean shift to select the number of clusters, while Mauro \textit{et al.}~\cite{7_mauro2013overlapping} employ the game theoretic model of dominant sets to find regular overlapping clusters. In a subsequent work~\cite{3_mauro2014integer}, Mauro \textit{et al.} propose to place selection after clustering and formulate an ILP problem with cameras as binary variables. The goal is to select the minimum number of views for each cluster, such that coverage and matchability are guaranteed. The same approach is adopted in this work. However, their formulation requires to execute the expensive Bron-Kerbosch algorithm~\cite{11_bron1973algorithm} for each keypoint in the cluster. Since neighboring keypoints are likely to share the same camera subgraph, a more efficient alternative is provided in Section~\ref{sect:selection}. 

Furthermore, all the methods presented so far cluster images according on their relative visibility information and without considering the 3D structure of the scene. Zhang \textit{et al.}~\cite{8_zhang2015joint} suggest to perform joint camera clustering and surface segmentation, which are formulated as a constrained energy minimization problem. Similarly, the clustering algorithm proposed in this work operates directly in 3D by exploiting the approximately uniform distribution of poses and geometry as produced by a moving vehicle.

It must be underlined that the naive solution of clustering images using temporal information from videos~\cite{10_akbarzadeh2006towards} is not ideal, since it does not consider multi-camera settings, where optimal viewpoints might be acquired by different sensors at very different time instants, and it fails when certain regions of the world are observed multiple times, since new clusters would be created each time, despite belonging to the same region.

Therefore, the proposed approach makes the following contribution to existing literature: (i) a novel framework that targets specifically urban scenarios and data acquired by a moving vehicle, which exhibit a unique set of challenges with respect to architectural Internet datasets; (ii) a clustering algorithm that is independent from pairwise relationships between cameras; (iii) a more efficient formulation of the ILP problem for view selection with respect to~\cite{3_mauro2014integer}.

\section{View Clustering\label{sect:clustering}}

\begin{figure*}
    \includegraphics[width = 0.99\textwidth]{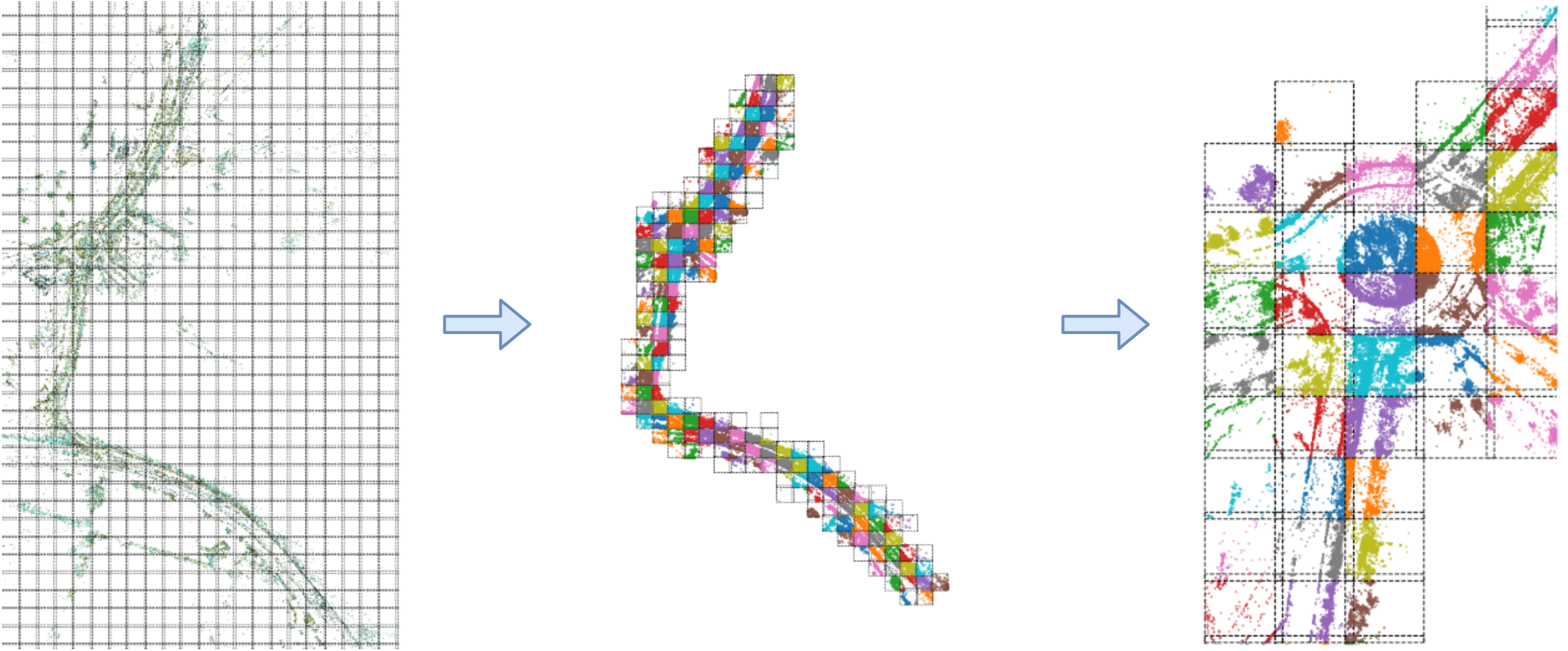}
    \caption{View clustering example: the raw 2D grid built from SFM (left), the full sequence clustered and filtered (center) and overlapping clusters in more detail (right). Each cluster is represented with its borders and in a different color.\label{fig:clustering}}
\end{figure*}

We assume images are acquired by a moving vehicle with a sensor suite of $N_{cams}$ cameras with framerate $f$ and processed by a SFM module to obtain intrinsic and extrinsic parameters, as well as a set of sparse keypoints, for each image. These data are the input to the view clustering algorithm, which starts by building a 2D grid on the $(x,y)$ plane with block size $(x_b,y_b)$ and overlap $d_{overlap}$, given the input range of the sparse keypoints. This process is shown in figure \ref{fig:clustering} (left): since the majority of clusters is empty and several blocks contain noisy keypoints very distant from the vehicle trajectory, a filtering step is required. To this end, each keypoint is assigned to the corresponding block simply by projecting it onto the $(x,y)$ plane and empty clusters without associated points are removed. In this way, the sparse input point cloud is divided into partially overlapping sets of keypoints which can be processed independently and in parallel. 

At this point, cameras must be associated to clusters. Intuitively, the assignment should be done based on the number and the quality of keypoints seen by each camera for each cluster. However, in urban scenarios with large textureless areas, the SFM module produces typically very few keypoints. Moreover, they are mostly concentrated in small textured patches of the world. Therefore, the 3D information of each non-empty cluster is augmented by sampling uniform points with resolution $r$ within the boundaries of the corresponding block. This allows cameras to be assigned in clusters within their field of view even if SFM did not extract any keypoint at that specific location.

Then, each set of points is projected onto each camera lying within a given distance from its centroid and all the cameras that see at least one point in the cluster are associated to it. Finally, clusters with a few number of cameras are iteratively merged with their neighbors, until a minimum target set of views is reached for each cluster. This lower bound is set to 10 views in experiments, in order to provide enough information for 3D reconstruction. The output of the view clustering algorithm is a set of of $C$ independent clusters of cameras with common visibility, shown in figure \ref{fig:clustering} (center) for a full sequence and in figure \ref{fig:clustering} (right) in a greater detail.

The key novelty of the proposed method is that it does not require the computation of complex pairwise relationships between cameras for the whole dataset, differently from previous literature~\cite{6_furukawa2010towards,3_mauro2014integer}. The use of the full similarity matrix quadratically scales as $O(N^2)$ and becomes quickly impractical in urban scenarios where up to $N = 60 \times f \times N_{cams}$ images are acquired every minute. Using the presented approach, a camera can be associated to at most $K$ neighboring clusters and this redundancy is independent from the size of the scene. In the worst case scenario, each cluster has $N_c = K \frac{N}{C}$ cameras associated. Even in this unlikely case, building $C$ separate and smaller visibility graphs is still significantly cheaper, requiring $O\left(\sum_{c=1}^{C} N_c^2\right)$ operations:

\begin{equation}
    \sum_{c=1}^{C} N_c^2 = \frac{K^2N^2}{C} < N^2 \Longleftrightarrow K < \sqrt{C}
    \label{eq:cond}
\end{equation}

In practical situations, a camera is associated at most to a cluster and its immediate neighbors (i.e. $K \leq 8$ by design), while the number of clusters grows quickly to several hundreds or thousands as the vehicle moves. This allows the approach to scale up to arbitrarily large scenarios, since the improvement gap grows as the dataset size increases. Furthermore, each block can be processed independently, thus boosting up the performance thanks to the high degree of parallelization that can be obtained. 

\section{View Selection\label{sect:selection}}

At the end of the view clustering algorithm, every camera is added to all clusters where associated points are present. This is likely to produce a highly redundant set of cameras for each block, since even a single point brings to associate a camera to a cluster. A view selection algorithm is then needed in order to choose the optimal subset of views for each cluster. In this context, \textit{optimal} means the smallest subset of cameras that guarantees that each point in the cluster is seen by at least $N_{vis}$ cameras and each camera have at least $N_{match}$ other cameras to be successfully matched with. 

Two cameras are considered to be \textit{matchable} if they see a sufficient number of common points. This differs from previous literature, where the typical similarity measure is the average Gaussian-weighted triangulation angle between the two camera centers and the common keypoints~\cite{6_furukawa2010towards,3_mauro2014integer}. In urban scenarios with very sparse features, this is not a reliable metric and it would require tuning the Gaussian parameters for each cluster, due to the high variability and sparsity of urban keypoints. Note that as the sampling resolution $r \rightarrow 0$, the number of common points is essentially a measure of the intersection between the two camera frustum and the cluster itself.

Therefore, an Integer Linear Programming (ILP) problem is formulated for each cluster, with binary variables $x_i \in \{0,1\}$ representing cameras:
\begin{equation}
\begin{array}{ll@{}ll}
\min  & \displaystyle\sum\limits_{i=1}^{N_{c}} x_{i} \\ [\bigskipamount]
\mathrm{s.t.} & \displaystyle\sum\limits_{i=1}^{N_{c}} x_{i} \geq N_{min} & \\ [\bigskipamount]
& \mathbf{A}_i^\top \mathbf{x} \geq 0 & \quad \forall i=1 ,\dots, N_{c} & \\ [\bigskipamount]
                 & \mathbf{B}_j^\top \mathbf{x} \geq N_{vis} & \quad \forall j = 1, \dots, P_c
                 
\end{array}
\end{equation}

The first constraint requires each camera in the cluster to have at least $N_{match}$ matchable cameras. A linear formulation can be derived from the similarity matrix $\mathbf{S}_c$ of the considered cluster as follows. Let $\tilde{\mathbf{S}}_c = \mathbf{S}_c > 0$ a binary matrix with $\tilde{s}_{c,ij} = 1$ if cameras $i$ and $j$ share common keypoints, $0$ otherwise. The constraint vectors $\mathbf{A}_i$ are computed as the rows of the matrix $\mathbf{A} = \tilde{\mathbf{S}}_c - N_{match} \cdot \mathbf{I}$, being $\mathbf{I}$ the identity matrix of size $N_c \times N_c$. This formulation effectively activates the constraint only for the selected cameras, ignoring the others.

The visibility constraint vectors $\mathbf{B}_j$ for each point in the cluster are composed by binary coefficients $b_{ij} = 1$ if the point $j$ is visible in camera $i$, $0$ otherwise. This is the first improvement with respect to the ILP formulation in~\cite{3_mauro2014integer}, where the coverage constraint requires that each point must be seen by at least one clique in the visibility subgraph associated to that point. This formulation implies that the Bron-Kerbosch algorithm for finding maximal cliques needs to be executed separately for each point. However, such requirement is both inefficient and redundant, since the same set of cameras is likely to see multiple points. 

As a second important difference, the efficiency of the algorithm is boosted further by providing a good initial guess to the ILP solver. A minimum number of cameras $N_{min}$ must be selected for each cluster, in order to guarantee enough information for the reconstruction. We select this threshold adaptively according the cluster size and clamp it between boundary values $N_{low}$ and $N_{high}$. Then, we sort cameras by the number of visible points and force the solver to select the $N_{min}$ views with the best visibility.

\section{Results\label{sect:results}}

\subsection{Experimental Setup}

The proposed framework has been implemented in C++ and tested on a consumer Intel Core i5-5300U 2.30 GHz CPU. The code relies on the open-source library OR-Tools~\cite{ortools} for solving the ILP problem during view selection. The approach is independent from the specific choice of both the SFM module producing the required input and the MVS software generating the dense reconstruction. Camera poses and sparse keypoints are obtained by a custom implementation of SFM with bundle adjustment, while the state-of-the-art MVS algorithm proposed in~\cite{13_xu2020planar} has been chosen, with the implementation provided by the authors.

To the best of our knowledge, the only two available large-scale urban datasets with a multi-camera setting are nuScenes~\cite{14_nuscenes2019} and the recently released DDAD~\cite{15_packnet}. However, they contain short sequences (20 s and 5-10 s, respectively) at a low framerate (12 Hz and 10 Hz, respectively), making it difficult to evaluate the proposed approach on those data. Therefore, the algorithm is validated on custom sequences acquired in the city of Parma, Italy, representing diverse real-world situations. The vehicle is equipped with $N_{cams} = 7$ cameras with resolution 3840 x 1920 and framerate $f = 30$~Hz.

The following set of parameters has been used for experiments: block size $(x_b, y_b) = (20, 20)$ m, with $d_{overlap} = 2$ m; sampling resolution $r = 1$ m; $N_{vis} = N_{match} = 2$, $N_{low}~=~10$ and $N_{high}~=~30$ for view selection.

\begin{table*}
\begin{center}
\begin{tabular}{|c|c|c|c|c|c|}
\hline
Dataset & Seq. 1 & Seq. 2 & Seq. 3 & Seq. 4 & Seq. 5\\
\hline\hline
\# views ($N$) & 5131 & 6782 & 8477 & 9912 & 11956 \\
\# keypoints ($P$) & 389621 & 321696 &  349616 & 372434 & 393246 \\
\# clusters ($C$) & 156 & 173 & 200 & 261 & 324 \\
\hline
$N$ after clustering  & 27103 & 32757 & 39014 & 49758 & 58364 \\
$K$ after clustering & 5.28 & 4.83 & 4.60 & 5.02 & 4.88 \\
Avg. $N_c$ after clustering & 173.75 & 189.34 & 195.07 & 190.64 & 180.13 \\
$t_{clustering}$ (s) & 22.35 & 28.52 & 32.7 & 37.32 & 40.89 \\
\hline
$N$ after selection  & 4951 & 5842 & 6266 & 8411 & 10138 \\
$K$ after selection & 0.96 & 0.86 & 0.74 & 0.84 & 0.85\\
Avg. $N_c$ after selection & 31.73 & 33.76 & 31.33 & 32.22 & 31.29 \\
$t_{selection}$ (s) & 50.54 & 76.13 & 92.4 & 124.85 & 144.02  \\
\hline
$t_{tot}$ (s) & 72.89 & 104.65 & 125.1 & 162.17 & 184.91\\
\hline
\end{tabular}
\end{center}
\caption{Quantitative results for each dataset and each stage of the pipeline.}
\label{tab:quant}
\end{table*}

\subsection{Quantitative Evaluation}

Table~\ref{tab:quant} provides a quantitative description of the input sequences and the algorithm results for each stage of the pipeline. The first thing to note is that urban data contain relatively few keypoints ($P$), when compared to architectural image sets with similar size~\cite{6_furukawa2010towards}. This justifies the choices of sampling additional points for each cluster and avoiding similarity measures based on the triangulation angle. Secondly, the clustering phase produces extremely redundant data ($N$ after clustering), with each view assigned to approximately 5 different clusters ($K$ after clustering), on average. Most of the clustering time ($t_{clustering}$) is spent for the camera association procedure, since several hundreds of views must be tested for each cluster. When processing extremely large sequences with millions of images, this phase can be naturally parallelized since each cluster is independent from the others, after the sampling step. Moreover, the optimization algorithm during view selection exploits the clustering redundancy to automatically remove useless views. The number of output views ($N$ after selection) is consistently lower than the input for each sequence, even considering that some cameras covering the overlap between two clusters are assigned to both. Finally, the efficiency condition imposed in equation~\eqref{eq:cond} ($K < \sqrt{C}$) is satisfied for each sequence by a large margin and this gap increases with the dataset size. 

\begin{figure}
    \centering
    \includegraphics[width=0.7\textwidth]{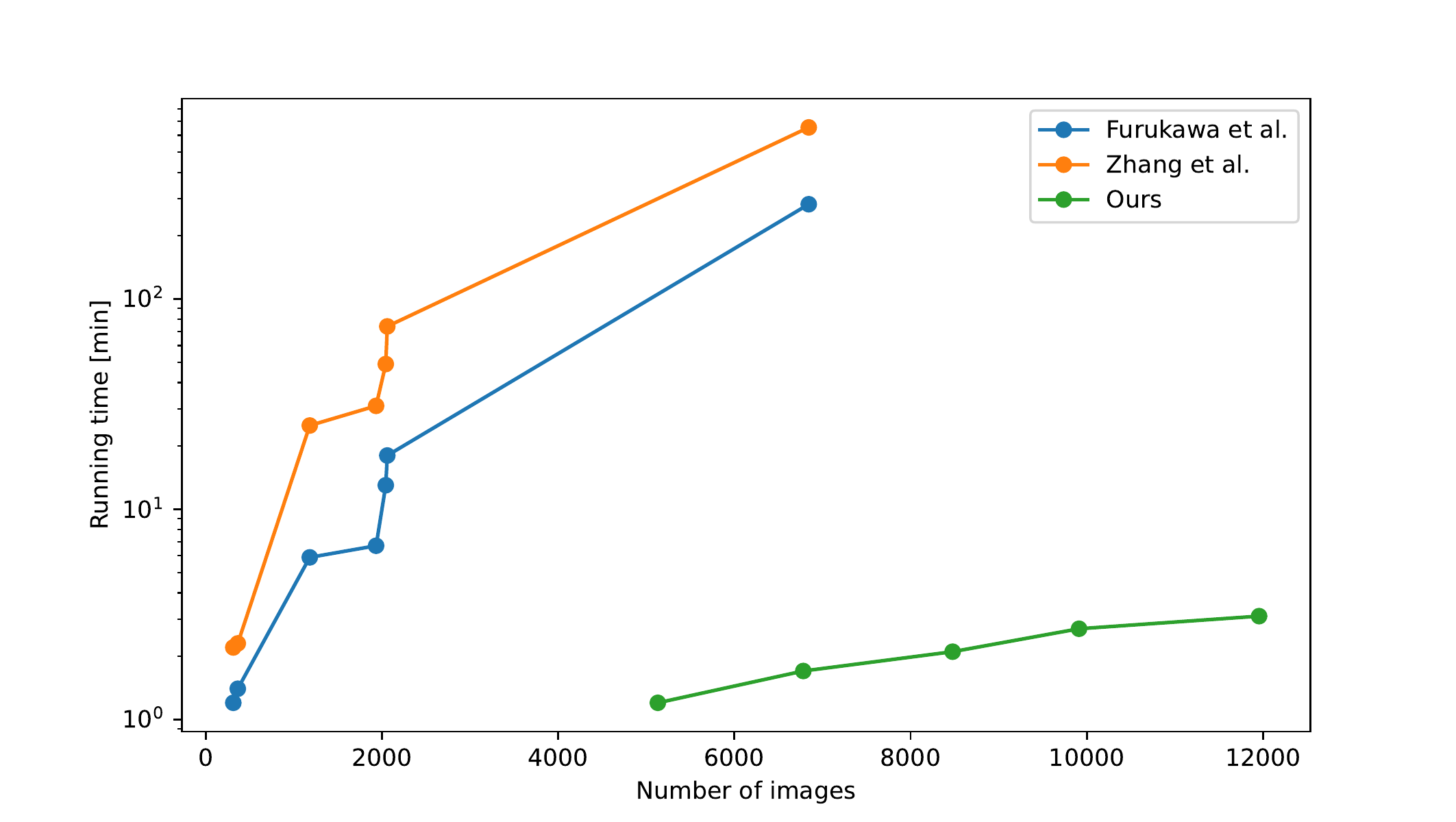}
    \caption{Running time comparison between our method and state-of-the-art solutions \cite{8_zhang2015joint,6_furukawa2010towards}. The Y axis is in log-scale for better visualization.}
    \label{fig:run}
\end{figure}

Direct quantitative comparison with state-of-the-art approaches is difficult, since they all target a substantially different scenario \cite{8_zhang2015joint,6_furukawa2010towards}, where the assumption of uniformly distributed poses is violated. In terms of the running time as a function of the input number of images, figure \ref{fig:run} shows that the proposed method is much faster and scales linearly with the dataset size, while \cite{8_zhang2015joint,6_furukawa2010towards} both require several hours to process a few thousands of images. This demonstrates that they are not suited for urban applications, where thousands of images are acquired every minute by the vehicle. The considered ILP baseline \cite{3_mauro2014integer} shares the same issue and numerical performances are available only for datasets smaller by an order of magnitude. The reported runtime for~706 images divided into 36 clusters is around 3~minutes. However, the Bron-Kerbosch algorithm scales exponentially as $O(3^{\frac{n}{3}})$ with the number of cameras $n$. In our collected sequences, an instance of such algorithm with $n \approx 10^2$ would be executed for each keypoint of each cluster, making the approach impractical even for very short scenes. On the other hand, our framework can process approximately 4000~images per minute.

\subsection{Qualitative Evaluation}

\begin{figure*}
    \begin{subfigure}{\textwidth}
      \centering
      \includegraphics[width=\linewidth]{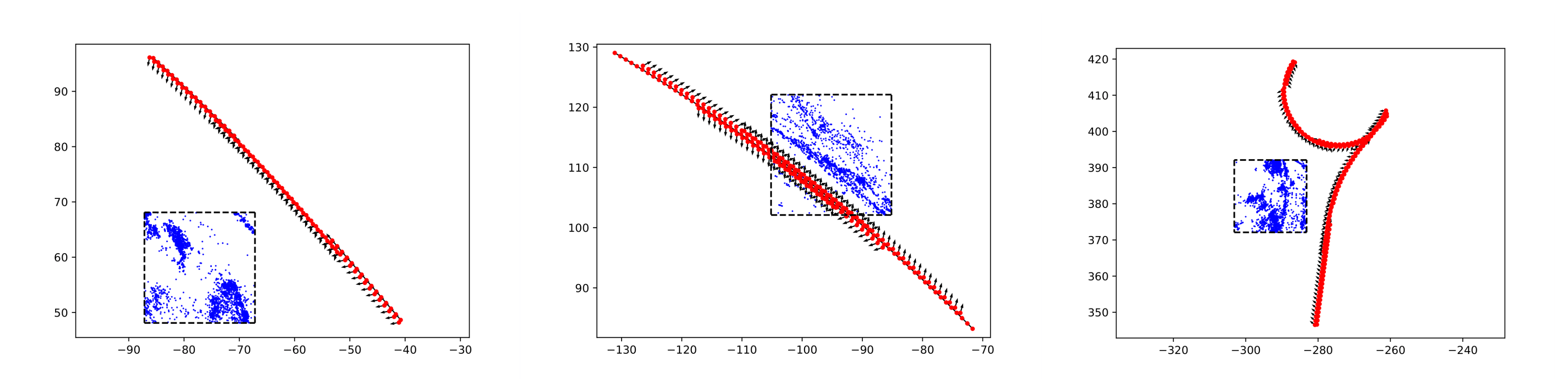}
      \caption{Clustering: basic (left), intersecting  (center) and complex (right). \label{fig:qual_clust}}
    \end{subfigure}
    \begin{subfigure}{\textwidth}
      \centering
      \includegraphics[width=\linewidth]{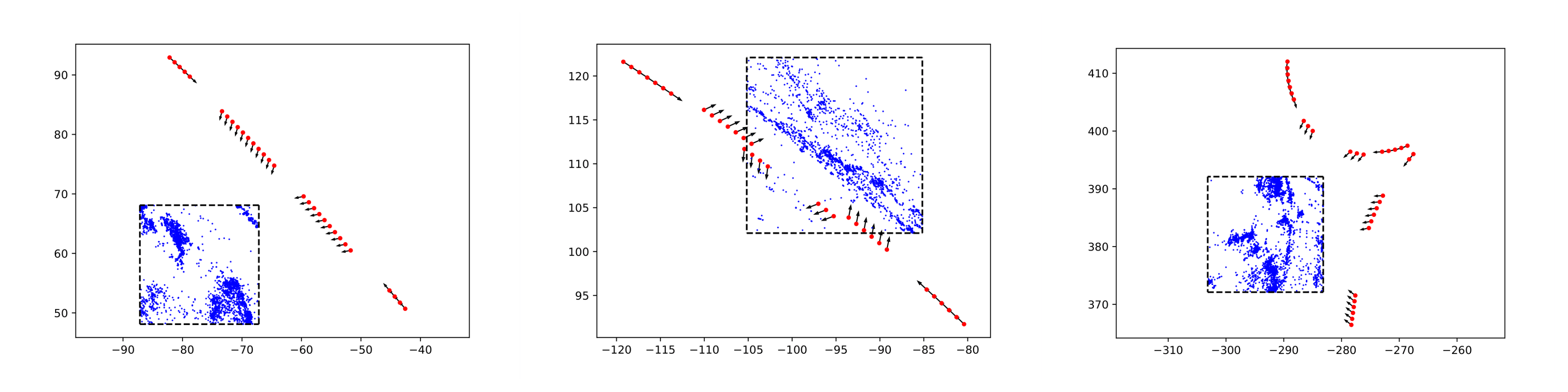}
      \caption{Selection: basic (left), intersecting  (center) and complex (right). \label{fig:qual_sel}}
    \end{subfigure}
    \caption{From clustering to selection: each cluster has black dashed borders, keypoints (blue) and cameras with viewing direction (red).\label{fig:results}}
\end{figure*}

Figure \ref{fig:results} provides a visualization of clustered cameras, before and after view selection, in order to show how redundancy is exploited and reduced by the optimization algorithm. Three main situations arise in urban data: (i) when all the views lie along a straight line outside the cluster boundaries (figure \ref{fig:qual_clust}, left), the ILP solver selects a well-distributed subset of cameras (figure \ref{fig:qual_sel}, left); (ii) when the vehicle trajectory intersects the cluster (figure \ref{fig:qual_clust}, center), two disjoint sets of cameras are selected (figure~\ref{fig:qual_sel}, center); (iii) for complex trajectories such as roundabouts (figure~\ref{fig:qual_clust}, right), the framework generalizes well by selecting cameras from diverse viewpoints (figure~\ref{fig:qual_sel}, right). 

\begin{figure*}
    \begin{subfigure}{\textwidth}
      \centering
      \includegraphics[width=\linewidth]{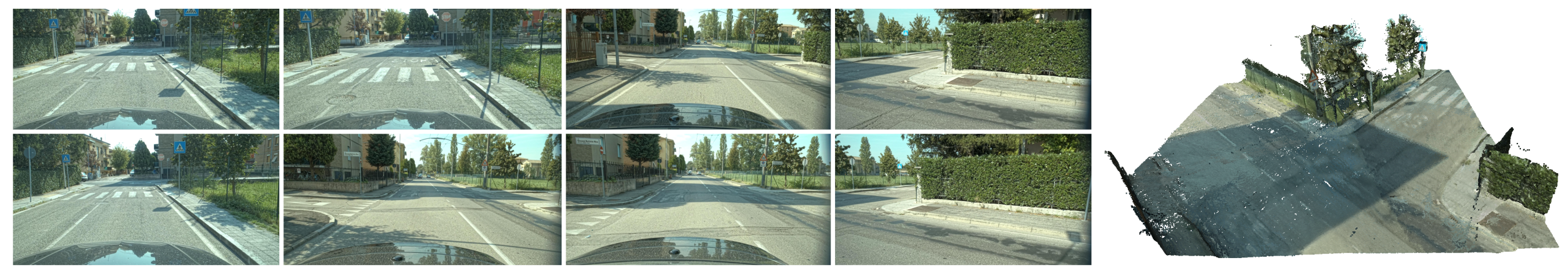}
    \end{subfigure}
    \begin{subfigure}{\textwidth}
      \centering
      \includegraphics[width=\linewidth]{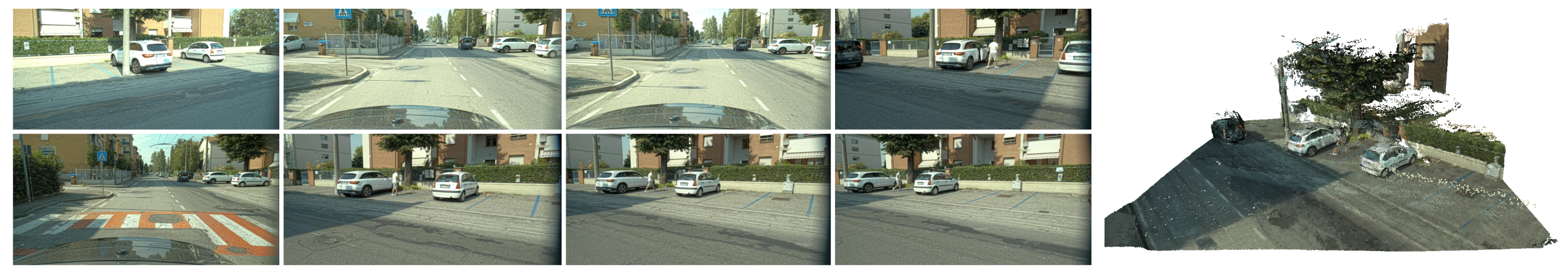}
    \end{subfigure}
    \begin{subfigure}{\textwidth}
      \centering
      \includegraphics[width=\linewidth]{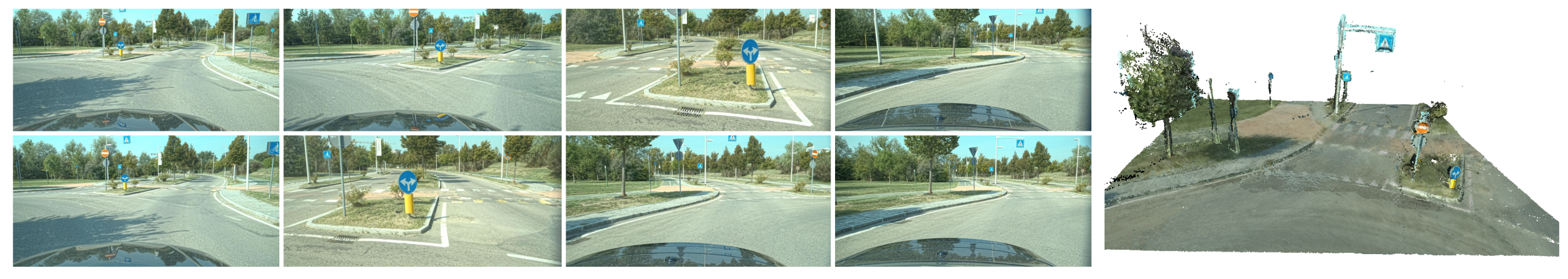}
    \end{subfigure}
    \caption{Clusters of images (left) and corresponding 3D point cloud (right).}
    \label{fig:3d}
\end{figure*}

Furthermore, figure~\ref{fig:3d} shows that the proposed method effectively cluster images based on shared visual content, which allows to produce a detailed 3D reconstruction of the world. Each point cloud has been cropped at the corresponding cluster boundaries and it is stored in this way for the subsequent global fusion step. Only qualitative evaluation of the resulting reconstruction is provided, since performances depend solely on the choice of MVS algorithm. The goal of the presented approach is to show that 3D reconstruction can be achieved in very large-scale scenarios where processing all the images in a single batch is not practically feasible. For a detailed comparison of state-of-the-art MVS algorithms for large-scale outdoor scenes, the reader can refer to~\cite{tanks}.

\begin{figure*}
    \centering
    \includegraphics[width=0.8\textwidth]{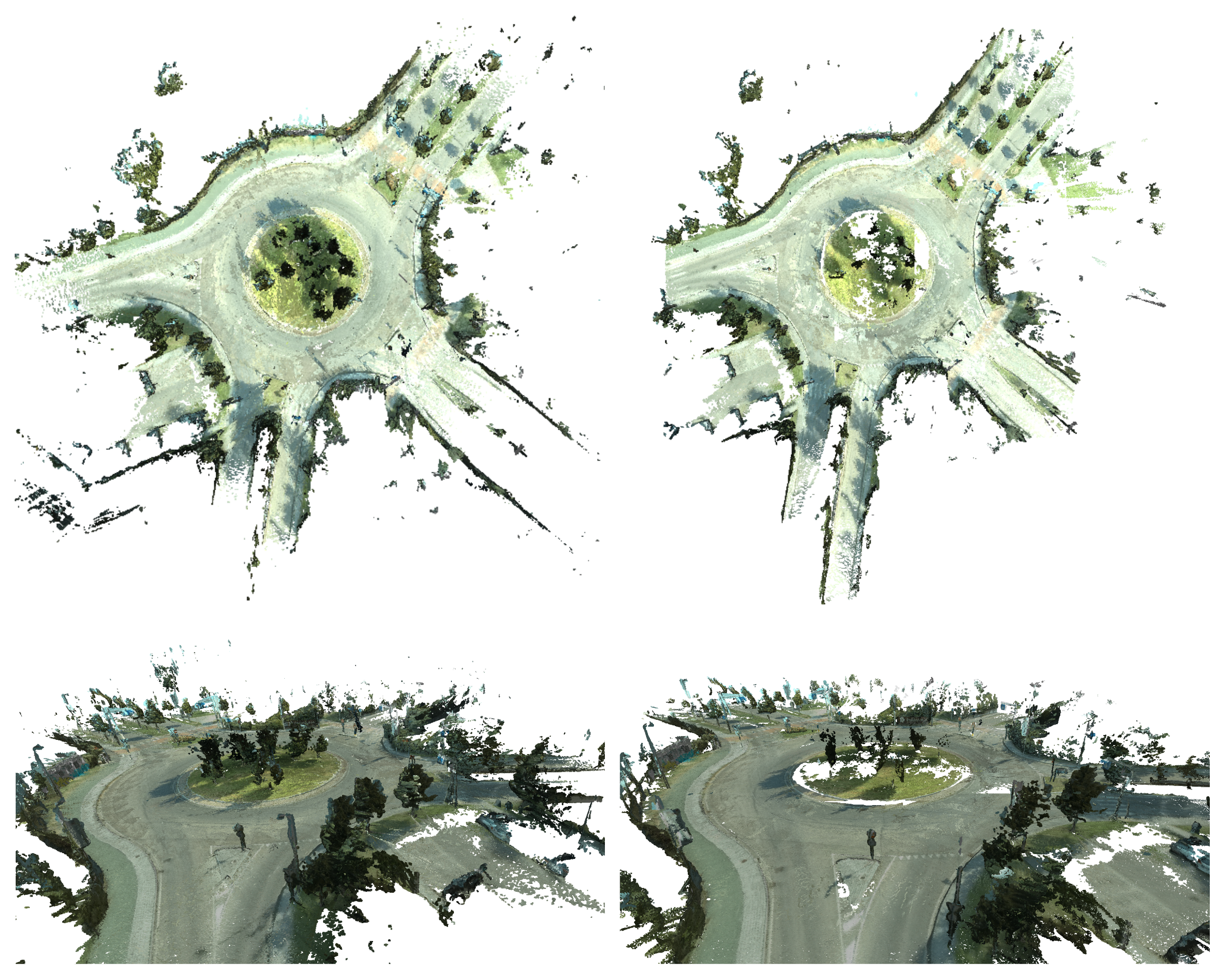}
    \caption{Point cloud comparison between batch reconstruction (left) and merged local clusters computed with the proposed algorithm (right).}
    \label{fig:batch}
\end{figure*}

Finally, a very short sequence with a few hundreds of images is considered, in order to have a relatively small instance of the problem where full reconstruction in a single batch is still possible. Figure~\ref{fig:batch} shows a comparison of the point clouds obtained by considering the whole set of images at once (left) and by merging multiple clusters computed with the proposed algorithm (right). While ground truth for numerical evaluation is not available, it can be seen that our divide-and-conquer approach maintains a good reconstruction quality, while being able to scale up to entire cities, where batch reconstruction is not an option.

\section{Conclusion\label{sect:conclu}}

In this work, a method for enabling city-scale 3D reconstruction from images acquired by a moving vehicle has been proposed. The algorithm builds a set of partially overlapping clusters over the vehicle trajectory, then it selects the optimal subset of views to compute a 3D point cloud, independently for each cluster. All the local point clouds are then fused together in order to obtain a full 3D model of the sequence. The presented framework focuses on efficiency, by clustering images independently from their pairwise similarities and providing a novel formulation for the view selection step. These contributions reduce the processing time with respect to state-of-the-art methods, not designed for the urban scenario, allowing the algorithm to scale up to arbitrarily large datasets.

%
%
%
%

\bibliographystyle{splncs04}
\bibliography{ref}

\begin{thebibliography}{10}
\providecommand{\url}[1]{\texttt{#1}}
\providecommand{\urlprefix}{URL }
\providecommand{\doi}[1]{https://doi.org/#1}

\bibitem{9_agarwal2011building}
Agarwal, S., Furukawa, Y., Snavely, N., Simon, I., Curless, B., Seitz, S.M.,
  Szeliski, R.: Building rome in a day. Communications of the ACM
  \textbf{54}(10),  105--112 (2011)

\bibitem{10_akbarzadeh2006towards}
Akbarzadeh, A., Frahm, J.M., Mordohai, P., Clipp, B., Engels, C., Gallup, D.,
  Merrell, P., Phelps, M., Sinha, S., Talton, B., et~al.: Towards urban 3d
  reconstruction from video. In: Third International Symposium on 3D Data
  Processing, Visualization, and Transmission (3DPVT'06). pp.~1--8. IEEE (2006)

\bibitem{11_bron1973algorithm}
Bron, C., Kerbosch, J.: Algorithm 457: finding all cliques of an undirected
  graph. Communications of the ACM  \textbf{16}(9),  575--577 (1973)

\bibitem{14_nuscenes2019}
Caesar, H., Bankiti, V., Lang, A.H., Vora, S., Liong, V.E., Xu, Q., Krishnan,
  A., Pan, Y., Baldan, G., Beijbom, O.: nuscenes: A multimodal dataset for
  autonomous driving. arXiv preprint arXiv:1903.11027  (2019)

\bibitem{6_furukawa2010towards}
Furukawa, Y., Curless, B., Seitz, S.M., Szeliski, R.: Towards internet-scale
  multi-view stereo. In: 2010 IEEE computer society conference on computer
  vision and pattern recognition. pp. 1434--1441. IEEE (2010)

\bibitem{15_packnet}
Guizilini, V., Ambrus, R., Pillai, S., Raventos, A., Gaidon, A.: 3d packing for
  self-supervised monocular depth estimation. In: IEEE Conference on Computer
  Vision and Pattern Recognition (CVPR) (2020)

\bibitem{tanks}
Knapitsch, A., Park, J., Zhou, Q.Y., Koltun, V.: Tanks and temples:
  Benchmarking large-scale scene reconstruction. ACM Transactions on Graphics
  \textbf{36}(4) (2017)

\bibitem{5_ladikos2009spectral}
Ladikos, A., Ilic, S., Navab, N.: Spectral camera clustering. In: 2009 IEEE
  12th International Conference on Computer Vision Workshops, ICCV Workshops.
  pp. 2080--2086. IEEE (2009)

\bibitem{3_mauro2014integer}
Mauro, M., Riemenschneider, H., Signoroni, A., Leonardi, R., Van~Gool, L.: An
  integer linear programming model for view selection on overlapping camera
  clusters. In: 2014 2nd International Conference on 3D Vision. vol.~1, pp.
  464--471. IEEE (2014)

\bibitem{4_mauro2014unified}
Mauro, M., Riemenschneider, H., Signoroni, A., Leonardi, R., Van~Gool, L.: A
  unified framework for content-aware view selection and planning through view
  importance. Proceedings BMVC 2014 pp. 1--11 (2014)

\bibitem{7_mauro2013overlapping}
Mauro, M., Riemenschneider, H., Van~Gool, L., Leonardi, R.: Overlapping camera
  clustering through dominant sets for scalable 3d reconstruction. Proceedings
  BMVC 2013  \textbf{2013},  1--11 (2013)

\bibitem{ortools}
Perron, L., Furnon, V.: Or-tools,
  \url{https://developers.google.com/optimization/}

\bibitem{1_schonberger2016structure}
Schonberger, J.L., Frahm, J.M.: Structure-from-motion revisited. In:
  Proceedings of the IEEE conference on computer vision and pattern
  recognition. pp. 4104--4113 (2016)

\bibitem{2_seitz2006comparison}
Seitz, S.M., Curless, B., Diebel, J., Scharstein, D., Szeliski, R.: A
  comparison and evaluation of multi-view stereo reconstruction algorithms. In:
  2006 IEEE computer society conference on computer vision and pattern
  recognition (CVPR'06). vol.~1, pp. 519--528. IEEE (2006)

\bibitem{13_xu2020planar}
Xu, Q., Tao, W.: Planar prior assisted patchmatch multi-view stereo. In:
  Proceedings of the AAAI Conference on Artificial Intelligence. vol.~34, pp.
  12516--12523 (2020)

\bibitem{8_zhang2015joint}
Zhang, R., Li, S., Fang, T., Zhu, S., Quan, L.: Joint camera clustering and
  surface segmentation for large-scale multi-view stereo. In: Proceedings of
  the IEEE International Conference on Computer Vision. pp. 2084--2092 (2015)

\end{thebibliography}

\end{document}